\documentclass[10pt,twocolumn,letterpaper]{article}

\usepackage{iccv}      

\definecolor{iccvblue}{rgb}{0.21,0.49,0.74}
\usepackage[pagebackref,breaklinks,colorlinks,allcolors=iccvblue]{hyperref}
\usepackage{booktabs}
\usepackage{multirow}
\usepackage{array}
\usepackage{pifont}
\usepackage{amsmath}
\usepackage{algorithm}
\usepackage{algpseudocode}
\usepackage{xcolor}
\usepackage{amssymb}
\usepackage{hyperref}
\usepackage{fancyhdr}
\usepackage{graphicx}  
\usepackage{url}
\newcommand\blfootnote[1]{%
\begingroup
\renewcommand\thefootnote{}\footnote{#1}%
\addtocounter{footnote}{-1}%
\endgroup
}
\renewcommand{\thefootnote}{\fnsymbol{footnote}}

\DeclareRobustCommand{\name}{InstructVEdit}

\definecolor{zccolor}{RGB}{0,0,0}
\newcommand{\zc}[1]{{\color{zccolor}#1}}
\definecolor{cjcolor}{RGB}{0,0,0}

\title{
InstructVEdit: A Holistic Approach for Instructional Video Editing
}

 \author{Chi Zhang\textsuperscript{1,*} \ 
Chengjian Feng\textsuperscript{2,*} \ Feng Yan\textsuperscript{2} \ 
Qiming Zhang\textsuperscript{3} \
Mingjin Zhang\textsuperscript{1} \\
Yujie Zhong\textsuperscript{2} \ Jing Zhang\textsuperscript{4,$\dagger$} \
Lin Ma\textsuperscript{2,$\dagger$} \\
\\
\textsuperscript{1}Xidian University \ \textsuperscript{2}Meituan Inc. \ \textsuperscript{3}University of Sydney \ \textsuperscript{4}Wuhan University
}

\begin{document}
\maketitle
\blfootnote{* Equal contribution. $\dagger$ Corresponding author.}

\begin{abstract}
Video editing according to instructions is a highly challenging task due to the difficulty in collecting large-scale, high-quality edited video pair data. This scarcity not only limits the availability of training data but also hinders the systematic exploration of model architectures and training strategies. While prior work has improved specific aspects of video editing (e.g., synthesizing a video dataset using image editing techniques or decomposed video editing training), a holistic framework addressing the above challenges remains underexplored. In this study, we introduce \name, a full-cycle instructional video editing approach that: (1) establishes a reliable dataset curation workflow to initialize training, (2) incorporates two model architectural improvements to enhance edit quality while preserving temporal consistency, and (3) proposes an iterative refinement strategy leveraging real-world data to enhance generalization and minimize train-test discrepancies. Extensive experiments show that \name achieves state-of-the-art performance in instruction-based video editing, demonstrating robust adaptability to diverse real-world scenarios. Project page: \url{https://o937-blip.github.io/InstructVEdit}.

\end{abstract}    
\section{Introduction}
\label{sec:intro}
With the rapid increase in the democratization of video creation, the demand for accessible and efficient video editing tools continues to grow. While powerful, conventional video editing tools often require specialized expertise and extensive manual effort. This barrier limits widespread creative participation, making advanced video editing inaccessible to the average user. To overcome these limitations, 
text-guided video editing has emerged as promising editing alternatives~\cite{wu2023tune,wu2024fairy,geyer2023tokenflow,ku2024anyv2v,khachatryan2023text2video,cheng2023consistent,singer2024video,polyak2025moviegencastmedia,wang2023zero,liu2024video,shin2024edit,qi2023fatezero,ceylan2023pix2video}. By leveraging natural language instructions, it enables users to edit videos with greater efficiency and precision, making video creation more intuitive, accessible, and inclusive.



Existing studies on text-guided video editing focus primarily on training-free methodologies~\cite{khachatryan2023text2video,ku2024anyv2v,geyer2023tokenflow,ceylan2023pix2video,kara2024rave}. Despite some advances, these approaches are hampered by restricted editing capabilities, suboptimal performance, and intricate pre-/post-processing necessities, thus hindering usability. Compared to training-free methods, training-based approaches offer significantly greater potential, as they enable control over broader aspects of video attributes rather than merely modifying limited features~\cite{cheng2023consistent,singer2024video,polyak2025moviegencastmedia}. However, collecting naturally supervised video editing data is highly impractical. 

\begin{figure}[tbp]
    \centering
    
    \begin{subfigure}[b]{\linewidth}
        \centering
        \includegraphics[width=\textwidth]{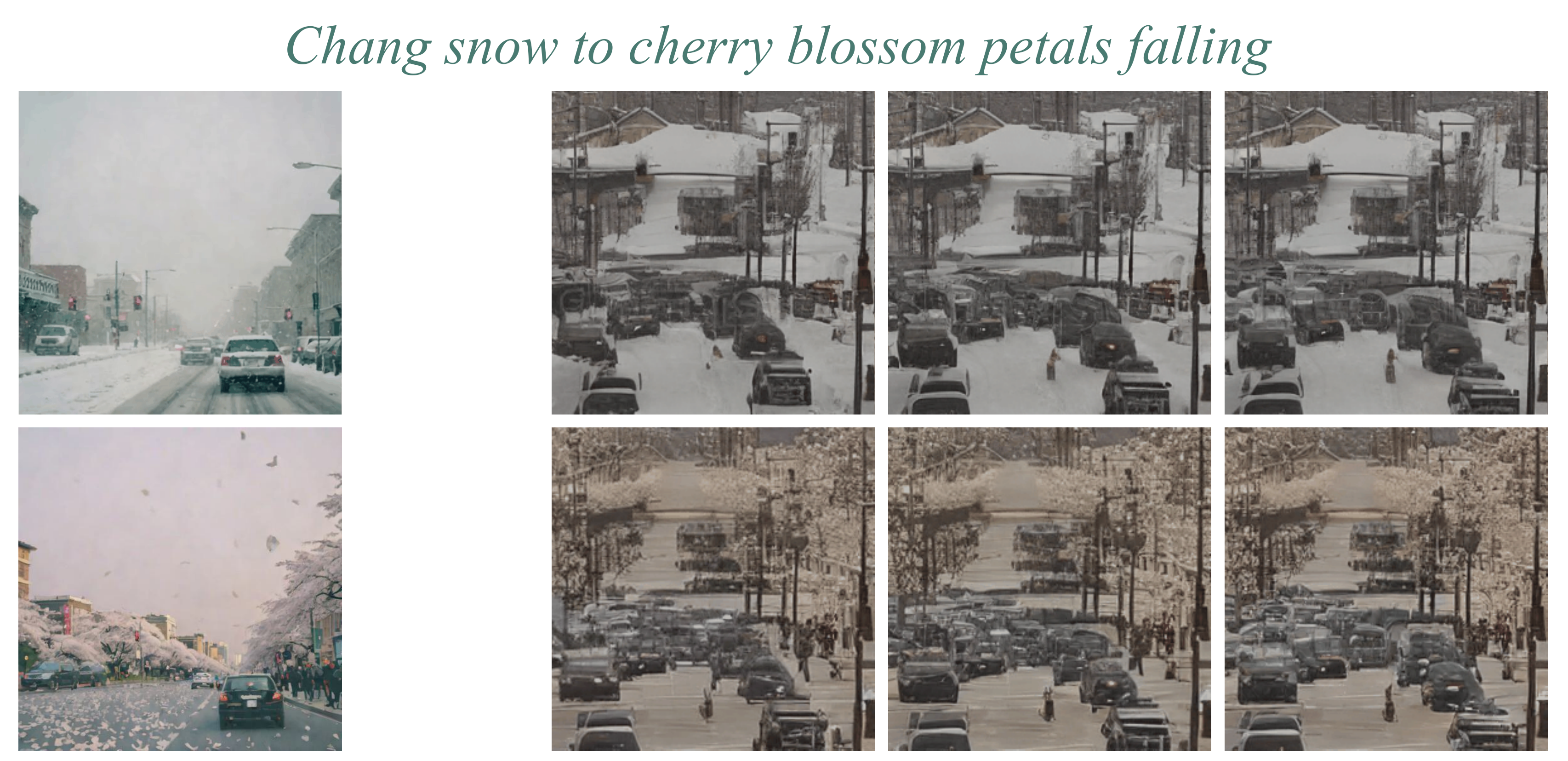}  
        \caption{Image editing pair and video editing pair generated by P2P~\cite{zhao2025ultraedit,cheng2023consistent}.}
        \label{fig:video_editing_comparison_p2p}
    \end{subfigure}
    \hfill
    \begin{subfigure}[b]{\linewidth}
        \centering
        \includegraphics[width=\textwidth]{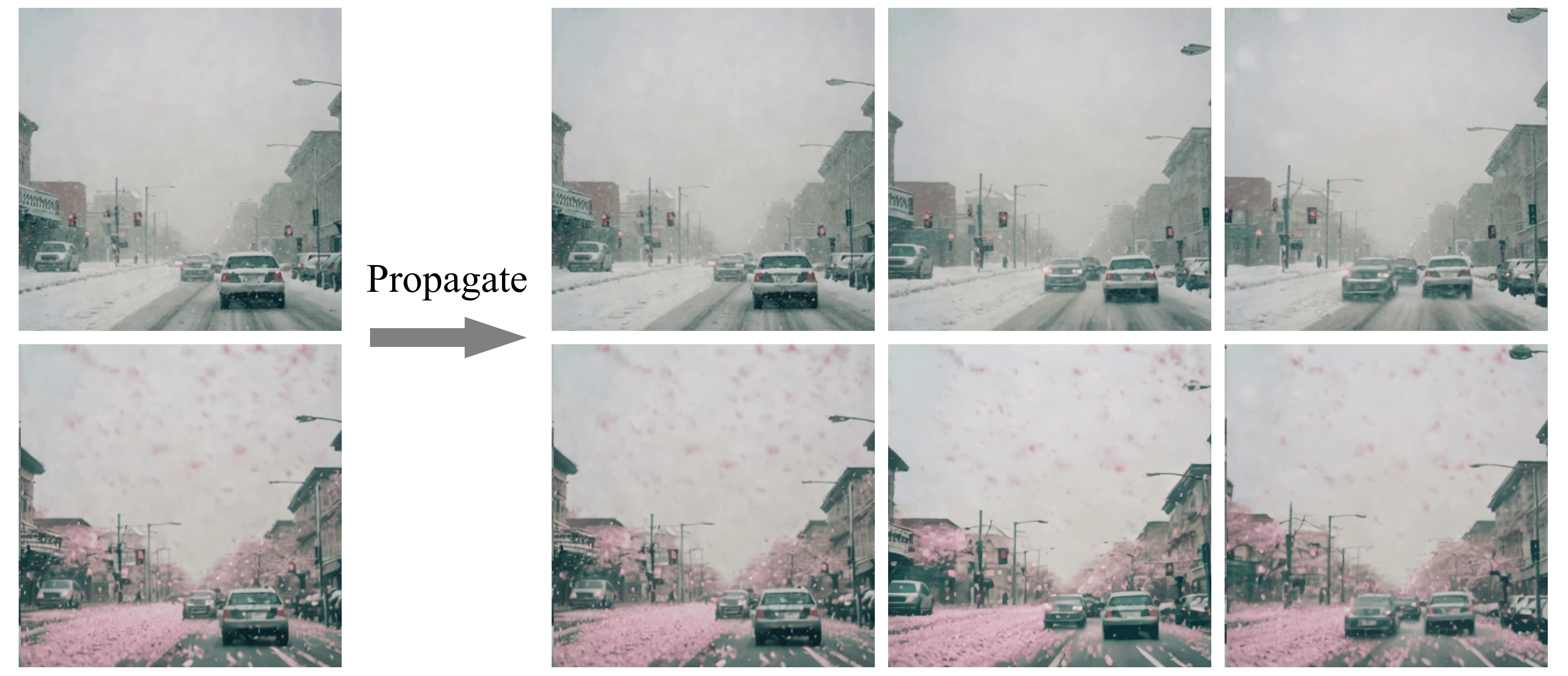}  
        \caption{Propagate the image editing pair to a video editing pair (\textbf{ours}).}
        \label{fig:video_editing_comparison_our}
    \end{subfigure}

    \caption{We compare our data curation approach with the existing method that directly utilizes P2P to generate video editing pairs.}
    \label{fig:video_editing_comparison}
\end{figure}

To overcome this limitation, InsV2V~\cite{cheng2023consistent} extends Prompt-to-Prompt (P2P)~\cite{hertz2022prompt} techniques—originally designed for image editing—to generate video editing data pairs. Nevertheless, applying P2P to videos often results in videos of lower quality compared to their image counterparts (See Fig.~\ref{fig:video_editing_comparison_p2p}), leading to suboptimal training data and offering limited improvements when used for model training. More broadly, the challenge of acquiring high-quality video editing supervision remains unresolved, 
limiting further exploration of effective model architectures.
For example, existing methods often resort to naively utilizing pre-trained modules from different sources and tasks, without considering their misalignment issue~\cite{cheng2023consistent,singer2024video}. Moreover, current training strategies face challenges in bridging the gap between training and real-world deployment, as they struggle to overcome the limitations imposed by suboptimal supervision data.

To address these issues, we propose a full-cycle pipeline, \name{} for instructional video editing, covering dataset curation, model design, and a multi-round iterative refinement strategy to enhance model performance.

First, we introduce a novel workflow to construct a well-curated video editing dataset. Technically, we start by training an image editing model using realistic-style image editing data. Next, we perform in-domain editing to acquire both the image pairs and their sampling trajectory, which is called \textbf{\textit{editing trajectory}} in our paper.
Then, we take the raw source image as the first frame and extend it to a source video clip by employing a pre-trained image-to-video model.
To generate the target video clips after editing, we develop a First-Frame-Guided Video-to-Video (FFG-V2V) model. It takes the editing trajectory of the initial frame within the attention module as guidance and translates source video clips to target ones frame by frame (including the first frame). This workflow enjoys high-quality and diverse editing capabilities from image editing. Besides, it also ensures temporal coherence at the same level as the source video by aligning the denoising direction of all target video frames with the editing trajectory of the initial frame pair, as shown in Fig.~\ref{fig:video_editing_comparison_our}. Unlike current training-free methods, which often rely on complex pre-processing, post-processing, multiple inference rounds in \zc{parallel} two-tower video generation models, and \zc{per-video fine-tuning}~\cite{ku2024anyv2v, geyer2023tokenflow, yatim2024space,ouyang2024i2vedit,wu2023tune}, our approach improves efficiency by streamlining the data preparation pipeline, reducing both computational load and time overhead as the model scale grows.

Secondly, we introduce two dedicated modules to make the image editing model a proficient video editing one. Specifically, we propose a soft motion adapter (SMA) to gradually inject motion priors into the image editing model. It helps the model hit a better Pareto frontier between temporal consistency and edit fidelity compared to previous methods (See Fig.~\ref{fig:sma_finetuned_sma}). Furthermore, we introduce an Editing-guided Propagation Module (EPM), which models spatiotemporal dependencies and manipulates the cross-frame relationship by dynamically adjusting the attention bias. The module ensures the frames with better edit effect exert a stronger influence over other frames during self-attention, rather than treating all frames equally.

Finally, to mitigate the synthetic-real discrepancy caused by training solely on synthetic data. We propose a multi-round iterative refinement strategy to continue training the generation model. We first collect real-world video datasets. Then for each round, we use these real-world video datasets to generate additional synthetic data by deploying the model trained in the previous round. The synthetic and real data construct video pairs, which are subsequently used to train a new video editing model in the current round.

Through multiple rounds of real-world video editing, filtering, and retraining, our approach progressively enhances the model's performance. As a result, the proposed model achieves state-of-the-art performance, demonstrates strong editing capabilities, and generalizes well to real-world scenarios.

To summarize, we make the following key contributions:
\begin{itemize}
    \item We present a comprehensive pipeline covering dataset curation, model design, and a multi-round iterative refinement strategy for instructional video editing. 
    \item We introduce a novel dataset construction workflow that delivers more reliable video pairs for model training, 
    significantly boosting model performance beyond previous dataset-based approaches.
    \item Building on the image editing model, we design a soft motion adapter (SMA) and an Editing-guided Propagation Model (EPM) that enhance the effectiveness of video editing.
    \item By employing a multi-round iterative refinement training strategy, we mitigate the synthetic-real discrepancy caused by training solely on synthetic datasets, delivering a model that achieves state-of-the-art performance.
\end{itemize}

\section{Related Work}
\label{sec:formatting}

\subsection{Diffusion models for text-based image editing}
Compared to text-based video editing, image editing based on specific instructions has a longer history~\cite{crowson2022vqgan,liu2020open,zhang2023adding,ruiz2023dreambooth,pan2023kosmos}. Recently, the emergence of large-scale diffusion models has significantly advanced this field~\cite{avrahami2022blended,kawar2023imagic,ramesh2022hierarchical,mokady2023null,meng2021sdedit}. For example, Prompt-to-Prompt~\cite{hertz2022prompt} modifies attention maps by injecting those from the input caption into the target ones. Such methods have greatly accelerated the development of constructing large-size image editing data~\cite{zhang2023magicbrush,sheynin2024emu,editinghq}. For instance, InstructPix2Pix (IP2P)~\cite{brooks2023instructpix2pix} uses Prompt-to-Prompt techniques to create large-scale image editing data. UltraEdit~\cite{zhao2025ultraedit} introduces a massive dataset anchored in real-world images. In our approach, we build our dataset and video editing model based on image editing, leveraging its broad editing capabilities and data availability from this field.

\subsection{Diffusion models for text-based video editing}
Text-based video editing has rapidly evolved with the emergence of diffusion-based models. One widely adopted approach to this task is the use of training-free solutions~\cite{wang2023zero,liu2024video,shin2024edit,qi2023fatezero,ceylan2023pix2video,khachatryan2023text2video,ku2024anyv2v,ouyang2024i2vedit}. For example,  TokenFlow~\cite{geyer2023tokenflow} achieves consistent video editing in a zero-shot manner by first editing keyframes and then propagating the changes to other frames. Tune-A-Video~\cite{wu2023tune} extends the self-attention mechanism to operate across multiple frames, ensuring that the generated frames maintain a globally consistent style. 

While training-free methods avoid computationally expensive optimization, their editing capabilities remain limited due to the restricted modification of input video features~\cite{polyak2025moviegencastmedia}. To enable more faithful and flexible editing, InsV2V~\cite{cheng2023consistent} tackles the scarcity of video editing datasets by extending the Prompt-to-Prompt technique to generate video data pairs. However, directly applying this technique for synthesizing video data yields suboptimal results, offering limited improvement when trained on such data. For the closed-source models, EVE~\cite{singer2024video} employs an unsupervised training approach through knowledge distillation. It leverages two expert models—one specialized in image editing and the other in text-to-video generation. However, the performance of this approach is inherently constrained by the capabilities of its two teacher models.

\section{Method}
In this section, we detail our \name{} method. Sec.~\ref{subsection: Problem Formulation} formulates the instruction-guided video editing problem. Sec.~\ref{subsection: Dataset Curation} introduces the data curation workflow, which serves as the foundation for initializing the model training. Sec.~\ref{subsection:model} presents the \name{} model architecture. Finally, Sec.~\ref{subsection:multiround} describes our multi-round refinement strategy for improving the model iteratively.

\subsection{Problem Formulation}
\label{subsection: Problem Formulation}
Instruction-guided video editing focuses on modifying a given video based on human instructions, such as `Make it snow' or `Turn the cat into a dog.' 
Let $V_s$ and $T_h$ denote the source video and human instruction, \emph{i.e.}, text, the video editing model $\Phi_\mathrm{v}$ aims to generate the target edited video:
\begin{equation}
    V_t = \Phi_\text{v}(V_s, T_h).
\end{equation}
This task presents significant challenges, not only due to the diverse range of editing scenarios—which span from local edits like object replacement and attribute manipulation to global modifications such as style transformation and background alteration—but more critically, due to the need to maintain temporal consistency across frames. 

In this paper, \textbf{we develop a comprehensive pipeline for building the video editing model}, \emph{i.e.}, we construct a high-quality video editing dataset $\mathcal{D} = \{(V_s, V_t, T_h)\}$ and propose a dedicated video editing model $\Phi_\mathrm{v}$, followed by an iterative training strategy to train the model for practical instruction-guided video editing.

\begin{figure*}[ht]
  \centering
    \includegraphics[width=\linewidth]{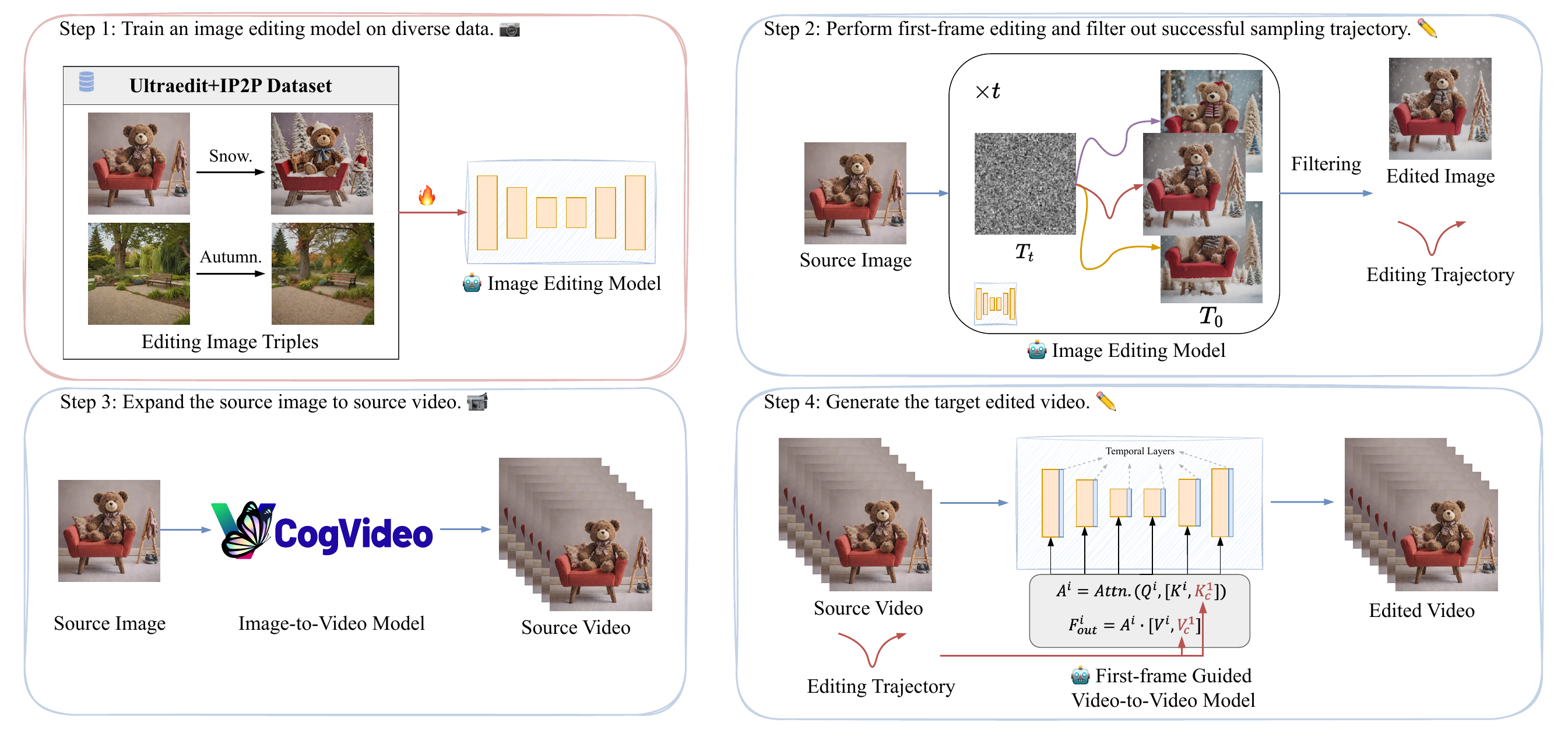}
   \caption{The data curation workflow. First, we train an image editing model primarily on a realistic-style dataset, then perform first-frame editing and filter out successful sampling trajectories. To construct video editing data pairs, we extend source images into source videos using an image-to-video model and generate the corresponding edited target videos with our proposed First-Frame-Guided Video-to-Video model. In our workflow, the training process is enclosed in a red frame, while the inference processes are enclosed in blue frames.}
   \label{fig:structure}
\end{figure*}

\subsection{Dataset Curation}
\label{subsection: Dataset Curation}
Here, we outline the curation process of the video editing dataset $\{(V_s, V_t, T_h)\}$. Unlike previous methods that directly generate synthetic video editing triples~\cite{cheng2023consistent}, our approach begins with image editing because instructional image editing is a well-established task with large-scale datasets that offer broad editing capabilities, making it a more reliable starting point for video data curation. Then, we extend the generated image to a video clip. As shown in Fig.~\ref{fig:structure}, our whole data creation process consists of four steps, with the first two steps aimed at generating image editing triple $(V_{s}^{1}, V_{t}^{1}, T_{h})$, where \(V_{s}^{1}\) and \(V_{t}^{1}\) represent the initial frame of a video clip that will be generated in the subsequent steps. The last two steps aim to convert the image editing triple into video triple $(V_{s}, V_{t}, T_{h})$.
Next, we detail the production process.

\textbf{Step 1: Training image editing model.} 
We utilize the representative IP2P model~\cite{brooks2023instructpix2pix} and train it with two datasets: UltraEdit~\cite{zhao2025ultraedit}, which focuses on real-world image editing, and P2P~\cite{brooks2023instructpix2pix}, known for its broad variety of editing prompts. By combining these datasets, we create a versatile model that excels in both realistic editing tasks and diverse transformations. In this work, our image editing model forms the foundation of the entire video editing pipeline.

\textbf{Step 2: Generating first-frame triple $(V_{s}^{1}, V_{t}^{1}, T_{h})$.}
To generate first-frame editing triples with a high success rate, we use the UltraEdit training data as \(V_{s}^{1}\) and run inference using the trained image editing model. Specifically, we follow IP2P~\cite{brooks2023instructpix2pix} and employ Classifier-Free Guidance~\cite{ho2022classifier} for image editing to generate diverse samples under varying text and image conditions. For each image, we perform 30 sampling iterations with 20 diffusion steps, varying the text guidance from 1.2 to 2.0 and the image guidance from 5.0 to 12.5 to explore a wide range of potential editing outcomes. \textit{We track the editing trajectory of the initial frame by storing its latent representations throughout the sampling process, which will inform the video editing in Step 4.}.

To refine the editing triples, we employ a two-stage filtering process. First, we follow UltraEdit~\cite{zhao2025ultraedit} to evaluate and filter the edited images using six automatic metrics, including DINOv2 similarity~\cite{oquab2023dinov2}, CLIP image similarity, SSIM, CLIP Directional Similarity~\cite{gal2022stylegan}, and CLIP similarity (as detailed in~\cite{zhao2025ultraedit}). Secondly, we normalize and assign weights to the six scores across all candidates to compute a total score. This allows us to select the final target image, $V_{t}^{1}$, for each source image, $V_{s}^{1}$. The weighting scheme is calibrated using human ratings to ensure that the automatic selection process aligns closely with human preferences.

\textbf{Step 3: Expanding $V_{s}^{1}$ to $V_s$.} 
In this step, we treat \(V_{s}^{1}\) as the initial frame and extend it into a high-quality video sequence, \(V_s\), using the powerful image-to-video generation model CogVideoX-5B-I2V~\cite{yang2024cogvideox}. Specifically, we utilize the selected images from Step 2, along with their corresponding captions from UltraEdit, as the image and textual conditions for CogVideoX-5B-I2V. This process generates a 48-frame video sequence.

\textbf{Step 4: Expanding $V_{t}^{1}$ to $V_t$.} 
To generate a high-quality edited target video \( V_t \) while ensuring temporal consistency, we propose the First-Frame-Guided Video-to-Video (FFG-V2V) model, which is built upon a pre-trained image editing model. The FFG-V2V model is used to translate the source video \( V_s \) into the target video \( V_t \) frame by frame, guided by the instruction \( T_h \). During the sampling process, FFG-V2V propagates the editing trajectory from the first frame \( V_s^1 \) across the entire video, by extending the self-attention module into a cross-frame attention module:
\begin{equation}
    F^{i}_{\mathrm{out}} = A^i \cdot [V^{i}, V^{1}_c],
\end{equation}
\begin{equation}
\resizebox{.90\linewidth}{!}{$
A^i = \mathrm{Attention}(Q^i,[K^i, K^1_c]) = \mathrm{Softmax}(\frac{Q^i [K^i, K^1_c]}{\sqrt{d}}).
$}
\end{equation}
Here $i$ denotes the $i$ frame of $V_s$. $Q$, $K$, and $V$ denote the query, key, and value, respectively. $d$ denotes the channel dimension of the query. \emph{$K^1_c$ and $V^1_c$ are the intermediate latent, i.e., key and value of the image $V_s^1$, which is obtained in Step 2.} By conditioning on the intermediate latent of a successful edit trajectory, the resulting video not only inherits high-quality edits but also guarantees uniformity in the editing direction across frames.

To further reinforce temporal consistency, we directly integrate the pre-trained temporal layers from AnimateDiff~\cite{guo2023animatediff} after each block of U-Net in FFG-V2V. These layers reshape the intermediate feature representation from $F \in \mathbb{R}^{b \times c \times f \times h \times w}$ to $F_{\mathrm{in}} \in \mathbb{R}^{(b \times h \times w) \times f \times c}$ and process it through the temporal module, enabling frame-wise temporal modeling:
\begin{equation}
    F_{\mathrm{out}} = F_{\mathrm{in}} + \mathrm{TemporalLayer}(F_{\mathrm{in}}).
\end{equation}
Thanks to the vast video data, these pre-trained temporal layers help improve motion consistency, even when they are not specifically trained to align with the other parts of the model, as observed in our experiments.

Next, we apply a filtering procedure similar to Step 2. For each video, we perform multiple sampling iterations, adjusting both text and image guidance to generate diverse stylistic results. These generated videos are then evaluated and filtered using five automatic metrics: ViCLIP$_{out}$~\cite{singer2024video}, ViCLIP$_{dir}$~\cite{singer2024video}, PickScore~\cite{kirstain2023pick}, ClipFrame~\cite{wu2023cvpr}, and ClipText~\cite{wu2023cvpr}. Details of these metrics are provided in Sec.~\ref{sec:exp}.

\begin{figure}[ht]
  \centering
    \includegraphics[width=\linewidth]{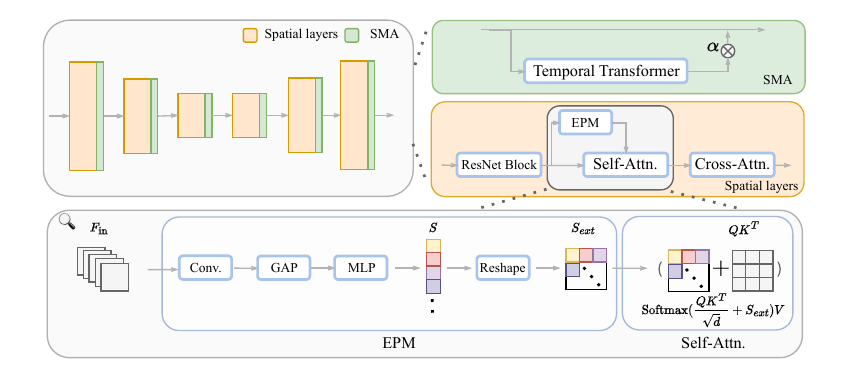}
   \caption{
   Overview of the \name{} model. We introduce two structural innovations: the Soft Motion Adapter (SMA) and the Editing-guided Propagation Module (EPM), which enhance the editing capabilities of the pre-trained image editing model while maintaining temporal consistency. The bottom section of the figure clarifies the information flow from EPM to self-attention.
   }
   \label{fig:model structure}
\end{figure}

\subsection{\name{} Model Design}
\label{subsection:model}

\subsubsection{Soft Motion Adapter (SMA)}
\label{subsection:Soft Motion Adapter}
Following the approach of previous methods~\cite{cheng2023consistent,singer2024video,wang2023zero}, we integrate temporal layers from AnimateDiff~\cite{guo2023animatediff} into our image editing model, enabling it to process video inputs. However, our experiments show a drop in the model's editing performance, even after fine-tuning.

\begin{figure}[tbp]
    \centering
    
    \begin{subfigure}[b]{\linewidth}
        \centering
        \includegraphics[width=\textwidth]{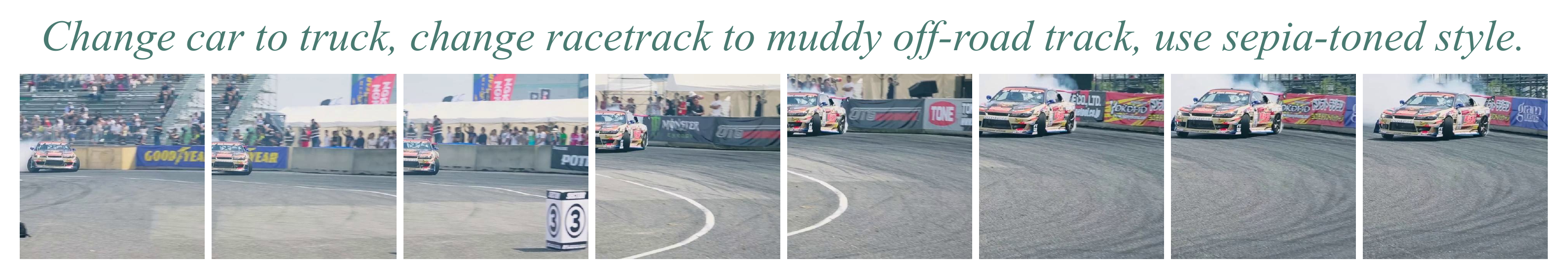}  
        \caption{Source video and editing prompt.}
        \label{fig:sma_source}
    \end{subfigure}
    \hfill
    \begin{subfigure}[b]{\linewidth}
        \centering
        \includegraphics[width=\textwidth]{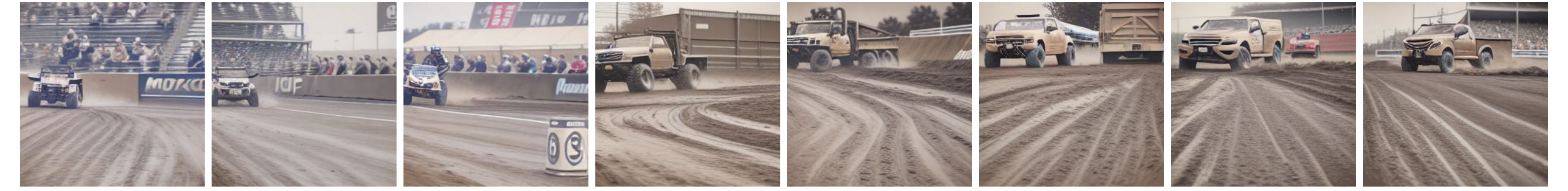}  
        \caption{Image editing model.}
        \label{fig:sma_image_edit}
    \end{subfigure}
    \hfill
    \begin{subfigure}[c]{\linewidth}
        \centering
        \includegraphics[width=\textwidth]{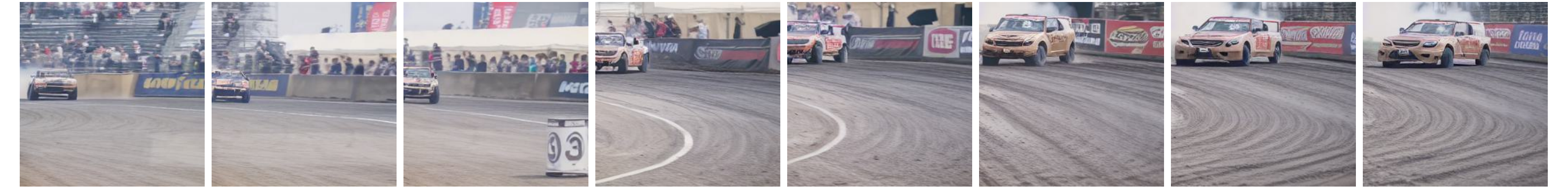}  
        \caption{Video editing model w/ pre-trained motion adapter.}
        \label{fig:sma_pretrained_ma}
    \end{subfigure}
    \hfill
    \begin{subfigure}[d]{\linewidth}
        \centering
        \includegraphics[width=\textwidth]{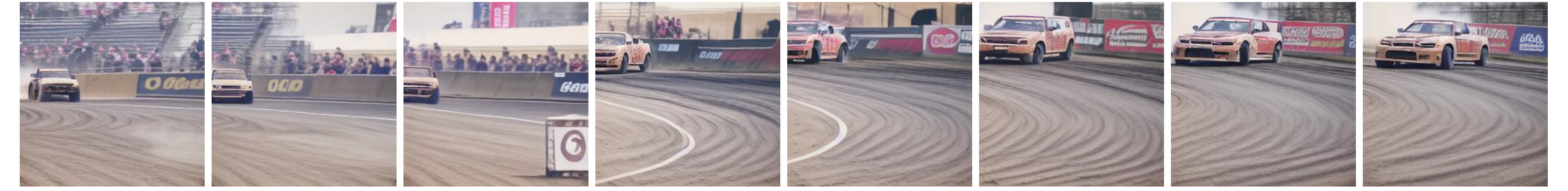}  
        \caption{Video editing model w/ fine-tuned motion adapter.}
        \label{fig:sma_finetuned_ma}
    \end{subfigure}
    \begin{subfigure}[d]{\linewidth}
        \centering
        \includegraphics[width=\textwidth]{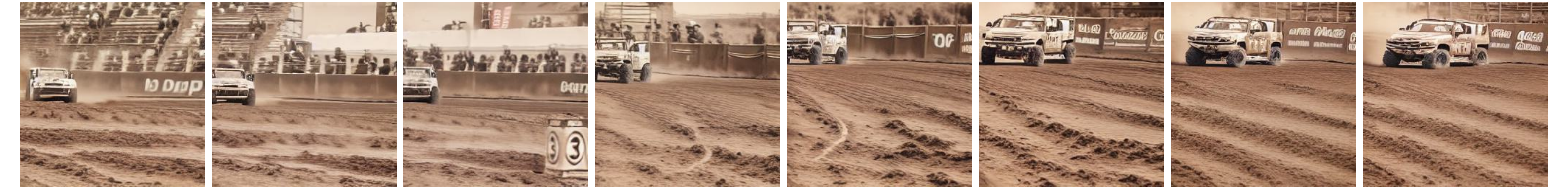}  
        \caption{Video editing model w/ fine-tuned SMA.}
        \label{fig:sma_finetuned_sma}
    \end{subfigure}
    \caption{Visualization of editing results from different models.}
    \label{fig:sma_and_epm}
\end{figure}

We hypothesize that this issue stems from the improper integration of the motion priors. To address this, we introduce a trainable parameter, $\alpha$, to soften each residual connection with temporal layers, dynamically adjusting its influence: 
\begin{equation}
    F_{\mathrm{out}} = F_{\mathrm{in}} + \alpha \times \mathrm{TemporalLayer}(F_{\mathrm{in}}),
\end{equation}
where $\alpha = 0$ at the beginning. As training progresses, the model gradually incorporates temporal information while largely retaining the editing ability of the base model, as illustrated in Fig.~\ref{fig:sma_finetuned_sma}. This simple yet effective design outperforms existing counterparts~\cite{cheng2023consistent,singer2024video}.

\subsubsection{Editing-guided Propagation Module (EPM)}
SMA effectively incorporates motion priors into the model, but it does not account for the quality of spatial edits. To address this, we propose the Editing-guided Propagation Module (EPM), as shown in Fig.~\ref{fig:model structure}, which jointly leverages both spatial and temporal cues. It captures bi-directional cross-frame dependencies and adaptively adjusts these relationships by modifying the attention bias. This approach ensures that better-edited frames have a stronger influence on other frames during self-attention, rather than treating all frames equally.

Given an input feature representation \( F_{\mathrm{in}} \in \mathbb{R}^{(b \times f) \times h \times w \times c} \), EPM first extracts spatial information from each frame using a two-layer \( 3 \times 3 \) convolution block, followed by global average pooling. This condenses the spatial data into a compact representation \( \mathbb{R}^{(b \times f) \times 1 \times 1 \times c} \), which is then reshaped to \( \mathbb{R}^{b \times (c \times f)} \). A two-layer MLP is applied to aggregate temporal information, predicting a matrix \( S \in \mathbb{R}^{b \times f^2} \) that captures bi-directional frame relationships. The process can be formulated as:
\begin{equation}
    S = \mathrm{MLP}(\mathrm{Reshape}(\mathrm{GAP}(\mathrm{Conv}(F_{\mathrm{in}})))).
    \label{equ:important}
\end{equation}

\begin{figure}[tbp]
    \centering
    
    \begin{subfigure}[b]{\linewidth}
        \centering
        \includegraphics[width=\textwidth]{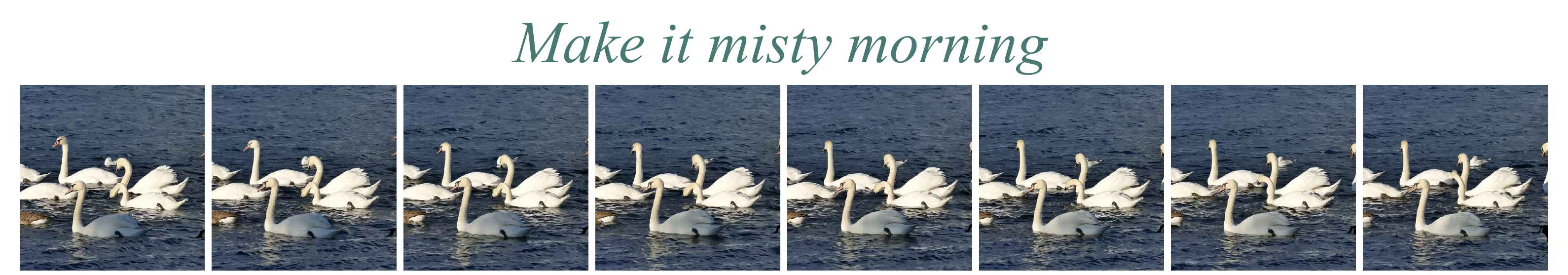}  
        \caption{Source video and editing prompt.}
    \end{subfigure}
    \hfill
    \begin{subfigure}[b]{\linewidth}
        \centering
        \includegraphics[width=\textwidth]{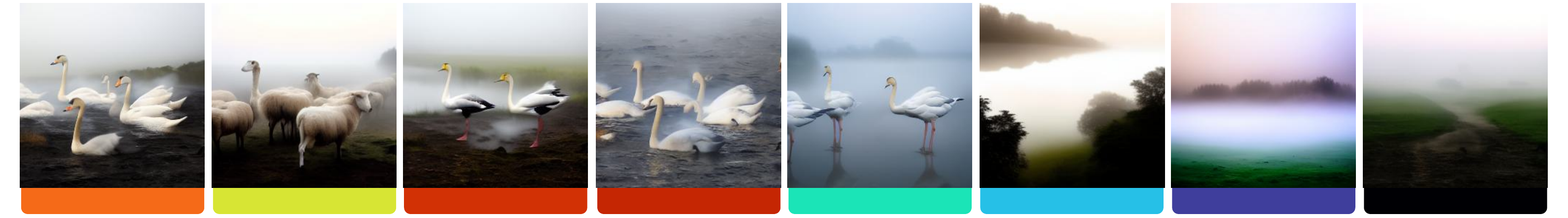}  
        \caption{Image editing model, and the heatmap of the importance sequence $s$.}
        \label{fig:ema_image_eidt}
    \end{subfigure}

    \begin{subfigure}[c]{\linewidth}
        \centering
        \includegraphics[width=\textwidth]{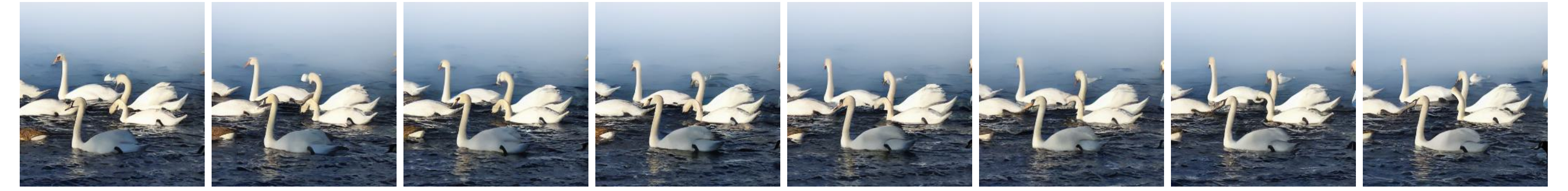}  
        \caption{Video editing model w/ EPM.}
    \end{subfigure}
    \caption{
    Visualization of editing results with and without EPM. To assess the impact of EPM, we perform inference on the source video (first row) while masking EPM's output. The importance sequence $s$ is computed as the column-wise average of $S$ from all EPM layers at the final denoising step, where each value represents the relative significance of a frame. The second row shows the resulting output and its heatmap, with red indicating higher importance and dark colors indicating lower importance. The third row illustrates the output with EPM enabled, highlighting frames with stronger edit effects and effectively propagating them.
    }
    
    \label{fig:epm}
\end{figure}

Then, we extend the self-attention module by integrating spatial and temporal tokens, transforming its input representation from $\mathbb{R}^{(b \times f) \times h \times w \times c}$ to $\mathbb{R}^{b \times (f \times h \times w) \times c}$. Next, we reshape and expand the importance matrix $S \in \mathbb{R}^{b \times f^2}$ to $S_{ext} \in \mathbb{R}^{b \times (f \times h \times w) \times (f \times h \times w)}$ and use it as an attention bias within the self-attention module to modulate token interactions.

Formally, the editing-guided cross-frame attention module can be formulated as:
\begin{equation}
    {A} = \mathrm{Softmax}(\frac{QK^T}{\sqrt{d}}+S_{ext}),
\end{equation}
\begin{equation}
    F_{\mathrm{out}} = A \cdot V.
\end{equation}

As illustrated in Fig.~\ref{fig:epm}, EPM identifies frames with optimal edit results and propagates them throughout the sequence to ensure effective editing and temporal consistency. At the beginning of training, we adjust the weight and bias of the MLP layer in Eq.~(\ref{equ:important}) by setting the diagonal elements of the attention bias $S_{ext}$ to 0 and the off-diagonal elements to -10. This initialization strategy helps maintain the pre-trained image editing model's original interaction patterns and avoids disrupting the learned representations.

\subsection{Multi-round Iterative Refinement}
\label{subsection:multiround}
Leveraging single-image editing, we generate realistic video triples to initialize model training. While these video triples significantly improve performance, both source and target videos are synthetic, creating a domain gap between training data and real-world videos.  To minimize this discrepancy, we propose an iterative refinement strategy that progressively enhances the model using real-world videos.

We begin by collecting source clips \( V_s \) from diverse real-world video datasets~\cite{vos2018,MOSE,hong2023lvos,MeViS}. Next, we generate corresponding edit instructions \( T_h \) using a multi-modal large language model (MLLM)~\cite{li2024llava}, fine-tuned on image editing datasets~\cite{zhao2025ultraedit,brooks2023instructpix2pix} for precise instruction generation.

Using these source videos and instructions, our video editing model generates the target video \(V_t\). Following Step 4 in Sec.~\ref{subsection: Dataset Curation}, we refine the generated videos. This approach helps minimize the domain gap between synthetic and real-world videos, as the source videos are sampled from real-world data and the generated target videos retain strong visual authenticity. Starting with the initial model from Sec.~\ref{subsection: Dataset Curation}, we iteratively enhance the generated videos and retrain the model through two rounds of refinement. This cyclical process continues until the model converges, progressively improving its real-world video editing capability.

\section{Experiments}
\label{sec:exp}

\begin{table*}[t]
    \centering
    \resizebox{\linewidth}{!}{
    \begin{tabular}{llccccc|cccc}
        \toprule
        Dataset & Method & ViCLIP$_{out}$$\uparrow$ & ViCLIP$_{dir}$$\uparrow$  & PickScore$\uparrow$ & CLIPFrame$\uparrow$&CLIPText$\uparrow$& Text & Struct. & Quality & Avg. \\
        \midrule
        \multirow{6}{*}{TGVE} 
        & TAV (CVPR'23)~\cite{wu2023tune} & 0.243& 0.162   &  20.36& 0.924 &  27.12 &  68.9 & 58.9 & 65.6 & 64.5\\
        & STDF (CVPR'24)~\cite{yatim2024space} & 0.226 & 0.110  &  20.40&  0.933 & 26.34 & 78.0 & 78.8 & 76.2 & 77.40 \\
        & Fairy$^\dagger$ (CVPR'24)~\cite{wu2024fairy} & 0.208& 0.164  &  19.80 &  0.933& / & / & / & / & / \\
        & TokenFlow (ICLR'24)~\cite{geyer2023tokenflow} & 0.257& 0.117  & 20.58& \textbf{0.943} & 25.41 & 70.0 & 49.6 & 60.0 & 59.9\\
        & AnyV2V (TMLR'24)~\cite{ku2024anyv2v}& 0.230 & 0.141 & 19.70 & 0.919 & 24.03 & 70.0 & 72.2 & 71.1 & 71.1\\
        & InsV2V (ICLR'24)~\cite{cheng2023consistent} & 0.262 & 0.208 &  20.76 &  0.911& 25.56 & 74.0 & 60.5 & 64.4 & 66.3 \\
        & EVE$^\dagger$ (ECCV'24)~\cite{singer2024video} & 0.262 & 0.221 &  20.76 &  0.922& / & / & / & / & /\\
        & Ours  & \textbf{0.280} & \textbf{0.237}  &  \textbf{20.92}&  0.919 & \textbf{27.69} & - & - & - \\
        \midrule
        \multirow{6}{*}{TGVE+} 
        & TAV (CVPR'23)~\cite{wu2023tune}& 0.242 &  0.131 &  20.47& 0.933 & 25.59 & 71.8 & 62.7 & 67.3 & 67.3 \\
        & STDF(CVPR'24)~\cite{yatim2024space}  & 0.227& 0.093  &  20.60& 0.933  & 25.70 & 74.7 & 76.4 & 77.0 & 76.0\\
        & Fairy$^\dagger$ (CVPR'24)~\cite{wu2024fairy}& 0.197& 0.140  &  19.81 & 0.933 & / & / & / & / & / \\
        & TokenFlow (ICLR'24)~\cite{geyer2023tokenflow}& 0.254 & 0.085 & 20.62 & \textbf{0.944} & 24.96  & 71.0 & 54.4 & 63.1 & 62.8 \\
        & AnyV2V (TMLR'24)~\cite{ku2024anyv2v}& 0.227 & 0.109 & 19.80 & 0.919 & 23.52 & 73.6 & 75.5 & 74.5 & 74.5 \\
        & InsV2V (ICLR'24)~\cite{cheng2023consistent}  & 0.236 & 0.174 &  20.37 & 0.925& 25.36 & 63.7 & 60.0 & 63.3 & 62.3 \\
        
        & EVE$^\dagger$ (ECCV'24)~\cite{singer2024video} & 0.251 & \textbf{0.198} &  20.88 & 0.926& / & / & / & / & / \\
        & Ours  & \textbf{0.271} & 0.183 & \textbf{20.94} & 0.917 & \textbf{26.65 }& - & - & - \\
        \bottomrule
        \multicolumn{4}{l}{\footnotesize $^\dagger$ Results from EVE~\cite{singer2024video}} 
    \end{tabular}}
    \caption{Performance comparison on TGVE~\cite{wu2023cvpr} and TGVE+~\cite{singer2024video} datasets. The best results are highlighted in bold.}
    \label{tab:results}
\end{table*}
\subsection{Experimental Settings}

\subsubsection{Metrics}
We employ automatic metrics and user studies to ensure a thorough evaluation. 
For automatic evaluation, following~\cite{cheng2023consistent,singer2024video,wu2023cvpr,polyak2025moviegencastmedia}, we utilize five metrics: 
1)~ViCLIP output similarity (ViCLIP$_{out}$~\cite{singer2024video}) – assesses the similarity between the edited image and output caption,
2)~ViCLIP text-video direction similarity (ViCLIP$_{dir}$~\cite{singer2024video} – evaluates the alignment between caption changes and video changes,
3)~Pickscore~\cite{kirstain2023pick} – measures average image-text alignment across video frames using a model aligned with human preferences, 4)~CLIPFrame~\cite{wu2023cvpr} - computes the average cosine similarity of CLIP image embeddings across frames, and 5)~CLIPText~\cite{wu2023cvpr} - quantifies average image-text alignment across frames.

We also conduct user studies for a more thorough evaluation, following the TGVE~\cite{wu2023cvpr} benchmark. In the study, each user is shown the input video, a caption describing the expected output, and two edited videos. They are asked to answer three questions: 1) Which video aligns better with the caption? 2) Which video better preserves the input video's structure? 3) Which video is more aesthetically appealing? An overall score is calculated by averaging the responses to all three questions.

\subsubsection{Datasets}
Our video editing pipeline comprises three stages: the first relies on real and synthetic images, while the second and third use real-world video data. \textit{Stage I}: We start by editing 50K images from the UltraEdit dataset~\cite{zhao2025ultraedit} to construct the synthesized image dataset. A selection process is then applied, filtering the results down to 25K high-quality images, which are then expanded into video sequences. After target video creation and filtering (Sec~\ref{subsection: Dataset Curation}), we obtain 9K videos meeting the required standards. \textit{Stage II}: We collect around 8K videos from diverse sources, including the YouTube-VOS~\cite{vos2018}, MOSE~\cite{MOSE}, LVOS~\cite{hong2023lvos}, and MeVis~\cite{MeViS} datasets. Then, we standardize them by clipping them into 16K 16-frame video segments. Our pre-trained model (trained on the synthesized dataset) is then applied for video editing, producing 3,402 video clips for training. \textit{Stage III}: We repeat the process from Stage II but use the updated model. This refinement iteration further improves the editing success rate, yielding 5,048 videos that meet the standard criteria for the final training phase.






\subsubsection{Video Editing Benchmark}
We conduct extensive experiments and analysis using the LOVEU-TGVE-2023 (TGVE)~\cite{wu2023cvpr} dataset from the CVPR'23 competition and the TGVE+ dataset proposed by EVE~\cite{singer2024video}. 
The TGVE dataset contains 76 videos, each accompanied by four editing prompts designed to evaluate different editing capabilities: style transformation, object replacement, background replacement, and multi-edit scenarios. The TGVE+ dataset expands on this by introducing three additional editing tasks: 1) object removal, 2) object addition, and 3) texture alterations.
For a fair comparison, we use the same editing instructions created by InsV2V~\cite{cheng2023consistent} and EVE~\cite{singer2024video}. Following their methodology, we apply a sliding window mechanism over the input video, generating 32-frame video sequences for evaluation.




\subsection{Main Results}
Here, we evaluate our model mainly training-based instructional video editing baselines, InsV2V~\cite{cheng2023consistent}, and EVE~\cite{singer2024video} and several popular training-free approaches including Tune-A-Video (TAV)~\cite{wu2023tune}, Space-Time Diffusion Features (STDF)~\cite{yatim2024space}, Fairy~\cite{wu2024fairy}, TokenFlow~\cite{geyer2023tokenflow}, and AnyV2V~\cite{ku2024anyv2v} for a comprehensive comparison. The qualitative results are shown in Tab.~\ref{tab:results}.

As shown in Tab.~\ref{tab:results}, \name{} achieves state-of-the-art performance across multiple evaluation metrics. Compared to training-based methods, our approach consistently outperforms InsV2V and EVE in ViCLIP\textsubscript{out}, ViCLIP\textsubscript{dir}, PickScore, and CLIPText, while maintaining a comparable CLIPFrame score. When compared to training-free methods, our approach also exhibits strong performance across most metrics, except for CLIPFrame. Notably, similar to other training-based methods, our method achieves a lower CLIPFrame score than training-free approaches. This is because CLIPFrame inherently favors static videos with minimal or no modifications, as noted by EVE~\cite{singer2024video}. 

\begin{figure*}[ht]
  \centering
    \includegraphics[width=\linewidth]{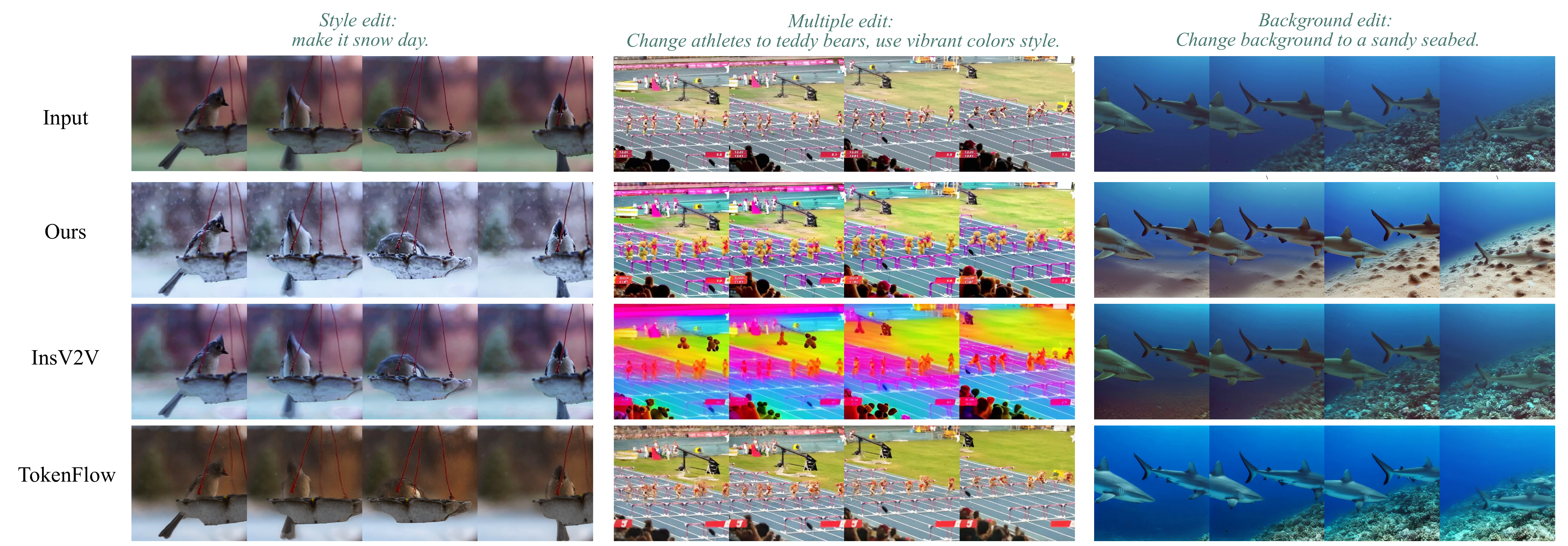}
   \caption{Comparison of our model against training-based and training-free method on samples from TGVE~\cite{wu2023cvpr} dataset.}
   \label{fig:visual_comparison}
\end{figure*}

Unlike the official TGVE contest, which primarily evaluates all participating methods against TAV, we follow EVE and directly benchmark our model against multiple baselines. According to our user study, participants consistently preferred our method in terms of text alignment, structural preservation, and aesthetic quality, demonstrating its superior ability to maintain edit fidelity and visual coherence in edited videos. Additionally, we provide qualitative comparisons between \name{} and baseline models in Fig.~\ref{fig:visual_comparison}.

\begin{table}[t]
    \centering
    \resizebox{\columnwidth}{!}{
    \begin{tabular}{lcccccc}
        \toprule
        Dataset & Data Volume & ViCLIP$_{dir}$$\uparrow$ & ViCLIP$_{out}$$\uparrow$ & CLIPFrame$\uparrow$ & PickScore$\uparrow$ & CLIPText$\uparrow$ \\
        \midrule
        InsV2V~\cite{cheng2023consistent} & 68k & 0.133 & 0.244 & \textbf{0.906} & 20.138 & 24.937\\
         Ours & 9k  &\textbf{0.167} & \textbf{0.251} & 0.902 &  \textbf{20.143}& \textbf{25.412} \\
        \bottomrule
    \end{tabular}
    }
    \caption{Comparison of our dataset against InsV2V~\cite{cheng2023consistent} dataset. The best values for each column are highlighted in bold.}
    \label{tab:dataset results}
\end{table}

\begin{table}[t]
\centering

\resizebox{\columnwidth}{!}{
\begin{tabular}{cc|ccccc}
 \toprule
SMA  & EPM  & ViCLIP$_{dir} \uparrow$ & ViCLIP$_{out} \uparrow$ & Pickscore $\uparrow$ & CLIPFrame$\uparrow$  & CLIPText$\uparrow$ \\
\midrule
\ding{55} & \ding{55}  &  0.167 & 0.251 & 20.143 & 0.902 & 25.412 \\
\ding{51} & \ding{55}  & 0.189 & 0.260 & 20.204 & 0.881 & 25.870 \\
\ding{55} & \ding{51}  & 0.187 & 0.260 & 20.375  & \textbf{0.903} & 26.282\\
\ding{51} & \ding{51}  & \textbf{0.194} & \textbf{0.263} & \textbf{20.407} & 0.899 & \textbf{26.654} \\
\bottomrule
\end{tabular}}
\caption{Ablation study results for the SMA and EPM. The experiments are conducted on the TGVE~\cite{wu2023cvpr} dataset.}
\label{tab:Model Design}
\end{table}

\begin{table}[t]
    \centering
    \resizebox{\columnwidth}{!}{
    \begin{tabular}{l c c c c c c }
        \toprule
        Stage & Data Volume &  ViCLIP$_{dir} \uparrow$ & ViCLIP$_{out} \uparrow$ & Pickscore $\uparrow$ & CLIPFrame$\uparrow$  & CLIPText$\uparrow$ \\
        \midrule
        \textit{Stage I} & 9k & 0.194 & 0.263 & 20.407& 0.899 & 26.654 \\
        \textit{Stage II} & 9k + 3.4k & 0.217 & 0.270 & 20.465 & 0.900 & 27.154 \\
        \textit{Stage III} & 9k + 5k & 0.227 & \textbf{0.272} & 20.441 & 0.898 & \textbf{27.279}   \\
        \textit{Stage III}$^\dagger$ & 5k & \textbf{0.228} & 0.271 & \textbf{20.504} & \textbf{0.901} & 27.227   \\
        \bottomrule
    \end{tabular}}
    \caption{Ablation study on iterative refinement. $^\dagger$ indicates training only on real-world data. The experiments are conducted on the TGVE~\cite{wu2023cvpr} dataset.}
    \label{tab:iterative}
\end{table}

\subsection{Ablation Study}
We conduct a series of ablation studies to analyze the contributions of different components on the TGVE dataset~\cite{wu2023cvpr}. To reduce computational cost, we use a fixed video classifier-free guidance~\cite{brooks2023instructpix2pix,cheng2023consistent} for all experiments. Our study focuses on three key aspects: synthetic dataset comparison, model design, and iterative refinement strategy. Below, we detail the experimental setup and results for each aspect.

\subsubsection{Synthetic Dataset}
To evaluate the effectiveness of our synthesized dataset, we follow the InsV2V~\cite{cheng2023consistent} setup and train identical models on both our dataset and the existing InsV2V dataset~\cite{brooks2023instructpix2pix}. As shown in Tab.~\ref{tab:dataset results}, InsV2V produces a much larger dataset (68K video pairs) than ours (9K video pairs). However, our dataset yields superior model performance.

Specifically, our model outperforms the one trained on the InsV2V dataset across ViCLIP$_{dir}$ (0.167 vs. 0.133), ViCLIP$_{out}$ (0.251 vs. 0.244), PickScore (20.143 vs. 20.138), and CLIPText (25.412 vs. 24.937), demonstrating stronger editing capability and improved text-video alignment. Despite being trained on a dataset nearly an order of magnitude smaller, our model achieves a comparable CLIPFrame score to InsV2V, indicating that our approach effectively preserves intra-frame coherence while requiring significantly less data.

\subsubsection{Model Design}
To better understand the contributions of different structural designs in our model, we perform ablation studies by selectively removing the SMA and EPM and evaluating their effects on performance. The results are summarized in Tab.~\ref{tab:Model Design}.

These experimental results confirm our earlier observation that the improper integration of pre-trained motion priors significantly weakens the model’s editing capability. Specifically, the baseline model, which excludes both SMA and EPM, maintains temporal consistency but struggles with effective editing, as reflected in its low ViCLIP$_{dir}$, ViCLIP$_{out}$, PickScore, and CLIPText scores (0.167, 0.251, 20.43, and 25.412, respectively). 
The introduction of SMA regulates the flow of motion priors into the model, largely preserving the raw editing ability of the image editing model. This improvement is evident in the higher ViCLIP$_{dir}$, ViCLIP$_{out}$, PickScore, and CLIPText scores. However, this enhancement comes at the cost of temporal inconsistency, as indicated by a lower CLIPFrame score. Meanwhile, introducing EPM alone enhances edit fidelity, visual quality, and frame consistency, as indicated by improvements across all metrics in Tab.~\ref{tab:Model Design}, demonstrating its effectiveness in spatial-temporal modeling.

Finally, with both SMA and EPM, our final model demonstrates significantly stronger editing capability compared to the baseline model, achieving: ViCLIP$_{dir}$ (0.194 vs. 0.167), ViCLIP$_{out}$ (0.263 vs. 0.251), PickScore (20.407 vs. 20.143), and CLIPText scores (26.654 vs. 25.412).

\subsubsection{Iterative Refinement}
Here, we provide the results after each stage in Tab.~\ref{tab:iterative} to ablate the effectiveness of our proposed pipeline. From \textit{Stage I} to \textit{Stage II}, most metrics show notable improvements. In particular, ViCLIP$_{dir}$ increases from 0.194 to 0.217 and CLIPText rises from 26.654 to 27.154, indicating better edit fidelity.

For \textit{Stage III}, we compare two training strategies: one trains on a combination of synthesized and real-world data (9k+5k samples) while the other trains exclusively on 5k real-world data. Interestingly, the 5k-only model achieves comparable or slightly better performance, suggesting that within this iterative refinement process, the final real-world data tends to surpass earlier synthetic data in quality. 
\section{Conclusion}
In this paper, we introduce \name{}, a holistic framework for instructional video editing that integrates data curation, model design, and iterative refinement. Our approach streamlines the generation of high-quality video editing pairs, enhances edit quality with SMA and EPM modules—ensuring motion consistency and effective spatial-temporal encoding—and iteratively incorporates real-world data to bridge the train-test domain gap. Extensive experiments demonstrate that \name{} achieves state-of-the-art performance, offering a robust and promising solution for instruction-based video editing.
 
{
    \small
    \bibliographystyle{ieeenat_fullname}
    \bibliography{main}
}

\end{document}